# The statistical trade-off between word order and word structure – large-scale evidence for the principle of least effort

Alexander Koplenig[1], Peter Meyer[1], Sascha Wolfer[1], Carolin Müller-Spitzer[1]

**1** Institute for the German Language (IDS), Mannheim, Germany.

## Abstract

Languages employ different strategies to transmit structural and grammatical information. While, for example, grammatical dependency relationships in sentences are mainly conveyed by the ordering of the words for languages like Mandarin Chinese, or Vietnamese, the word ordering is much less restricted for languages such as Inupiatun or Quechua, as those languages (also) use the internal structure of words (e.g. inflectional morphology) to mark grammatical relationships in a sentence. Based on a quantitative analysis of more than 1,500 unique translations of different books of the Bible in almost 1,200 different languages that are spoken as a native language by approximately 6 billion people (more than 80% of the world population), we present large-scale evidence for a statistical trade-off between the amount of information conveyed by the ordering of words and the amount of information conveyed by internal word structure: languages that rely more strongly on word order information tend to rely less on word structure information and vice versa. In addition, we find that – despite differences in the way information is

expressed – there is also evidence for a trade-off between different books of the biblical canon that recurs with little variation across languages: the more informative the word order of the book, the less informative its word structure and vice versa. We argue that this might suggest that, on the one hand, languages encode information in very different (but efficient) ways. On the other hand, content-related and stylistic features are statistically encoded in very similar ways.

## Introduction

Natural languages employ different strategies to transmit information that is necessary to recover specific aspects of the corresponding message (e.g. grammatical relations, thematic roles, agreement phenomena, and more generally, the encoding of grammatical categories). While, for example, grammatical information ("who did what to whom") in a sentence is mainly conveyed by the ordering of the words in languages like Mandarin Chinese or Vietnamese, the word ordering is much less restricted for languages like Inupiatun or Quechua, as those languages (also) use the internal structure of words (e.g. the modification of word roots by inflection or the compounding of roots) as cues to inform about grammatical relationships in a sentence. This has led linguists to speculate, mostly qualitative in nature [1–5], about a potential trade-off between the amount of regularity of the ordering of words and the amount of regularity of the internal word structure: languages that rely more on word order to encode information rely less on morphological information and vice versa. In this paper, we explicitly address this question quantitatively.

Theoretically, the trade-off hypothesis can be justified as an instantiation of Zipf's principle of least effort [6], or the more general framework of synergetic linguistics [7]: If, for example, grammatical relationships in a sentence are fully determined by the ordering of words, it would constitute unnecessary cognitive effort to additionally encode this information by intra-lexical regularities. If, however, word ordering gives rise to some extent of grammatical ambiguity, we should expect this ambiguity to be cleared up with the help of word structure regularities in order to avoid unsuccessful transmission. If we define $D_{structure}$ as the amount to which parts of a word token can be predicted given the observed overall regularities in intra-word structure and $D_{order}$ as the amount of information that is expressed in the ordering of words [8], then the simplest conceivable mathematical form of this relationship would be:

$$D_{structure} * D_{order} = c \qquad [1]$$

where $c$ is some constant. Put differently, we set out to test the hypothesis that $D_{order}$ varies inversely as $D_{structure}$. Or, if we relax the rather strong assumption of a proportional relationship, we can model $D_{order}$ as a function of the reciprocal of $D_{structure}$, where the conditional expectation can be written as:

$$E(D_{structure}|D_{order}) = \beta_0 + \beta_1 * (D_{order})^{-1} \qquad [2]$$

where the parameters $\beta_0$ and $\beta_1$ are estimated empirically. If we do not want to make any assumptions regarding the functional form of the relationship, we can compute Spearman's rank correlation coefficient $r_s$ between $D_{order}$ and $D_{structure}$. A trade-off between both variables would imply that:

$$r_s \ll 0 \qquad [3]$$

## Material and Methods

We neither pursue liturgical or theological goals, nor do we want to propagate Christian missionary work. As linguists, our interest in the Bible stems solely from the fact that it is the book with the most available translations into different languages [9].

Interactive visualizations, raw data and code to reproduce all results presented in this paper, are available online at http://www.owid.de/plus/eebib2016/project.html. On a side note, our data also conform to a typical areal pattern of linguistic features, viz. "a strong tendency to geographical homogeneity" [2]. The online presentation of our results makes such patterns accessible through interactive maps.

As our data basis, we used the Parallel Bible Corpus made available by [9]. It contains 1,559 unique translations of the Bible in 1,196 different languages in a fine-grained parallel structure (regarding book, chapter and verse). Each translation is tokenized and Unicode normalized. Spaces were inserted between words and both punctuation marks and non-alphabetic symbols. In addition, all texts were manually checked and corrected by [9] where necessary. In texts without spaces or marks between words, a dictionary lookup method was used to detect word boundaries (e.g. for Khmer, Burmese, or Mandarin Chinese). Detected word tokens are space-separated. All uppercase characters were lowered in a language-specific way based on the closest ISO 639-3 code provided by [9]. We then split each bible translation into different books of the biblical canon, effectively treating each book as a different text sample of the corresponding Bible translation. Here, we focused on the following six books of the New Testament: the four Gospels (Matthew, Mark, Luke, John), the Book of Acts and the Book of Revelation, because (i) we have enough available translations in different languages for those books and (ii) those books are reasonably long which makes the estimation of our two key quantities more reliable and robust. Interactive visualizations for all other books of the biblical canon are available online.

In October 2015, the Wycliffe Global Alliance estimated that almost 6 billion people have access to at least portions of the Bible in their native language [10]. To check the reliability of this figure, we extracted native speaker estimates for the languages available to us from the English Wikipedia [https://en.wikipedia.org, all accessed on 07/11/2016]. Our estimate corresponds well with the figure quoted above. It is important to emphasize that such a figure has to be treated with caution, because (i) census methods and dates of surveying vary significantly, (ii) defining a language, a language variety, or a dialect can be difficult, and (iii) there are people with more than one native language.

Let us represent each book as a symbolic sequence of $N$ characters i.e. $b = \{c_1, c_2, ..., c_{N-1}, c_N\}$ where $c_i$ represents any character (including white spaces and punctuation marks) in the book at position $i$. The set of all distinct characters (or letters) that appear in $b$ is defined as the alphabet $A^b$, while the set of all distinct space-separated sequences (or word types) that appear in $b$ is defined as the book's lexicon $W^b$. While this technical definition of word types is the de-facto standard in quantitative linguistics, this definition can be called into question from a theoretical point of view as mentioned above [11,12].

Since we are interested in measuring the amount of information that is conveyed by the word structure and the ordering of words, we use information theory as our mathematical framework [13]. To this end, we use one of the key ideas of the Minimum Description Length Principle: "any regularity in the data can be used to *compress* the data, i.e. to describe it using fewer symbols than needed to describe the data literally. The more regularities there are, the more the data can be compressed." [14]. We measure the entropy per symbol or entropy rate $H^b$ for each book $b$ which can be defined [13] as the average amount of information that is needed in order to describe $b$. Or put differently, $H^b$ measures the redundancy of $b$ [15]. We use the non-parametric estimation method of [15,16] that is based on the Lempel-Ziv compression algorithm [17]. This method does not require any prior training, produces robust estimates without the need for very long strings as input and is able to take into account very long range correlations typical of literary texts [18,19] that are not captured by direct parametric Markovian or "plug-in" estimators [15]. For each book $b$, we estimate the per-symbol description length as [15]:

$$\widehat{H}^b = \left[\frac{1}{N}\sum_{i=1}^{N}\frac{l_i}{\log(i+1)}\right]^{-1} \qquad [4]$$

To measure the minimum number in bits per character [bpc], logarithms throughout this paper are taken to base two. Here, the key quantity of interest is the match-length $l_i$. It measures the length of the shortest substring starting at position $i$ of $b$ that is *not* also a substring of the part of the book before this position. For example, given the string "montana bananas" at position $i = 10$, $l_{10}$ is equal to 4 ("anan"). Roughly speaking, at position $i$, the past $\{c_1, ..., c_{i-1}\}$ characters of the sequence are taken as a database (or lexicon) to assess how much of the sequence starting at $i$ is contained somewhere in our "database". The intuitive idea of this approach is that longer match-lengths are indicative of more redundancy in the source texts and, therefore, a lower mean uncertainty per letter. There are no restrictions regarding the size of the "database", illustrating (i) why the estimator can be used in the presence of very long range correlations, as we do not impose any restrictions on how far "into the past we can look for a long match" [15] and (ii) that the estimator seems like a reasonable model of linguistic patterns of experience, as it captures structure on various levels of linguistic organization (co-occurring words, regular relations between grammatical word forms, constructions) that can be linked to theories of language learning and language processing [20]. Details of our Java implementation and an open source version can be found online.

It is important to note that the estimation of entropy rate is defined as the average description length for a process that is both *stationary* and *ergodic*. It is not clear *a priori* if textual data can be seen as such a process [13], or if both concepts are even meaningful for natural languages [21]. To induce (at least some) stationarity, and thereby improve convergence, we simply randomized the order of the verses in each book, effectively discarding all supra-verse information [22].

Based on the ideas of [8,23,24], we approximate the amount of information that is conveyed by ordering of words and the structure of words, by estimating $\widehat{H}_b$ for three versions of each book: (i) $\widehat{H}^b_{original}$ is estimated on the basis of the original version of the book. (ii) $\widehat{H}^b_{order}$ is estimated the basis of a version of the book where word ordering has been deliberately destroyed. (iii) $\widehat{H}^b_{structure}$ is estimated on the basis of a version of the book in which intra-lexical regularities have been masked. Now, if we use the book version with absent word order to construct a code instead, we will need $\widehat{H}^b_{original} + \widehat{D}^b_{order}$ *bpc* on average in order to describe *b* where $\widehat{D}^b_{order}$ is the relative entropy [13,24]. Analogously, $\widehat{D}^b_{structure} = \widehat{H}^b_{structure} - \widehat{H}^b_{original}$. Thus, the incurring penalty of $\widehat{D}^b_{order}$ or $\widehat{D}^b_{structure}$ measures the amount of information in *bpc* that gets lost on average if the ordering of words or the intra-lexical structure is not considered when an efficient code is constructed in order to compress *b*. Hence, higher value of $\widehat{D}^b_{order}$ or $\widehat{D}^b_{structure}$ are indicative of a greater amount of regularity or information of the word order or the word structure.

Table 1 illustrates our approach of (i) destroying the word order and (ii) masking the word structure. For the version of the book with absent word order, we simply randomized the order of words within each verse. This means that when estimating the entropy rate of this book, redundancy that stems from the word order cannot be used to compress the corpus, but the statistics on the word level remain constant. In languages where the relative ordering of words is free, this manipulation should not have a major influence. Hence, a small $\widehat{D}^b_{order}$ indicates that the relative ordering of words is less informative in the respective language. For the version of the book with masked word structure information, all tokens for each word type $w_j \in W^b$ with a length of at least 2 characters are replaced by a unique equal-length sequence of characters randomly constructed from the alphabet $A^b$, effectively destroying the structure on the word level, but keeping both the syntactical and the collocational structure constant. For example, Table 1 shows that the intra-lexical regularity of forming the simple past tense via the suffix "ed" is masked as well as the regularity of forming nouns from stems of Latin via the suffix "tion". If the intra-lexical structure carries less information in a particular language, meaning that most words have little or no internal structure, this manipulation should not have a major influence on the entropy rate and thus lead to a small $\widehat{D}^b_{structure}$.

| Original | **i** said **i** just dropped **in** to see what **condition** my **condition** was **in** |
| Masked word structure | **i** nypa **i** wpid imjnwct **nc** ye wuj jaoh **hywjopoic** ue **hywjopoic** wea **nc** |
| Destroyed word order | **condition** said was **i in** what dropped **i** my see to **in condition** just |

Table 1: Toy example illustrating our approach. **First line**: original text (one line taken from Kenny Rogers' song "Just Dropped In"). **Second line**: masked word structure. [*NB.*: "i" is not masked since it is only one character long. Thus it does not contain any intra-lexical structure.] **Third line**: destroyed word order. Word types printed in boldface appear two times.

It is worth pointing out that in both manipulated versions, basic quantitative structural properties of the original text remain unaffected (e.g. book length, word length, the type-token word frequency distribution), which rules out the (likely) possibility that changes to those characteristics influence the entropy rate estimation [25]. For the inter-book comparisons (cf. Figure 4 & Figure 5), we keep $N$ constant by first identifying the book with the smallest size in characters and then truncating the other five books at this position. In 1,450 of all 1,476 translations with available information for the six books, the shortest book is Revelation. Since we randomized the order of words within each verse for the book version with absent word order, differences of (average) verse lengths (in words) between the six books could potentially influence our results. To rule out this possibility, we generated an additional data set, where $N$ is kept constant, but the word order is randomized per book instead of verse. The results based on this data set are qualitatively indistinguishable from the results we report here. The data set is avaiblable online.

To understand the functional form of the relationship between $\widehat{D}^b_{structure}$ and $\widehat{D}^b_{order}$ (cf. Eq. 2 & Figure 1), we fitted the following non-linear regression function by least squares for each book *b*:

$$\widehat{D}^{b,t}_{structure} = \beta_0 + \beta_1 * (\widehat{D}^{b,t}_{order})^{-1} + \epsilon^{b,t} \quad [5]$$

where $t = 1, 2, \ldots, T$ are our available translations for $b$; $\epsilon^{b,t}$ is the error term. For languages with more than one available translation $\widehat{D}^b_{structure}$ and $\widehat{D}^b_{order}$ were averaged across languages, except when stated otherwise in the text or in Figure 5 where we used all translations with available information for all six books.

*p*-values for the correlation of the rankings of the six books for the selected languages are based on exact permutation tests for all $6! = 720$ permutations.

## Results

Figure 1 summarizes the primary result of our analysis. There is clear evidence for a statistical trade-off between the amount of word structure information and the amount of word order information. For all investigated books, there is a negative Spearman correlation of at least $r_s = -.71$.

While our results do not permit the conclusion that $D_{order}$ is inversely proportional to $D_{structure}$ in a strictly mathematical sense (cf. Eq.1), the black dashed lines in Figure 1 show that a lot of the structure can be captured by a simple statistical model that suggests the presence of a reciprocal relationship (cf. Eq. 2): variation in $D_{order}$ explains at least 54% of the variation in $D_{structure}$ for all six investigated books. Interactive visualizations for all other books of the biblical canon are available online.

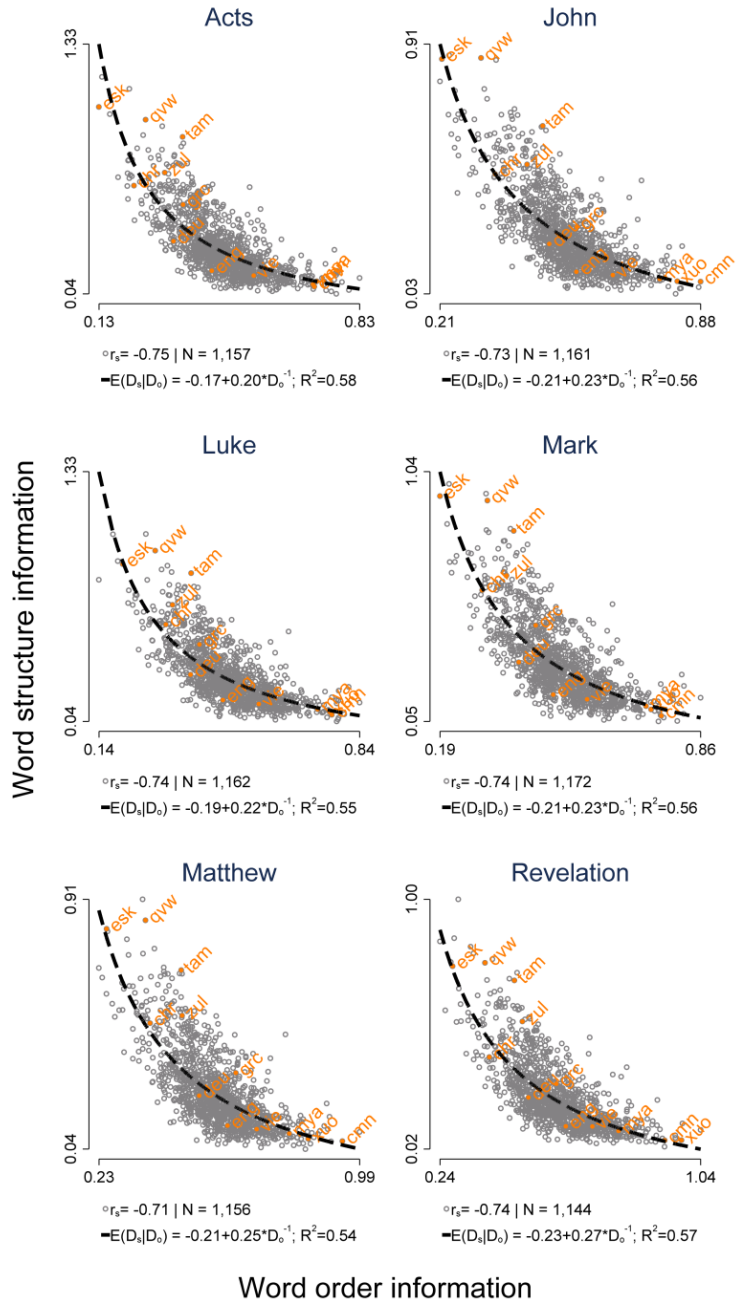

Figure 1. The statistical relationship between word structure information and word order information (in *bpc*) for six investigated books of the biblical canon. Orange labels show the ISO codes for twelve selected languages. $r_s$-values are Spearman correlation coefficients. Black dashed lines in each plot indicate that a lot of structure can be captured in a simple model that suggests the presence of a reciprocal relationship (cf. Eq.2). Abbreviations: chr – Cherokee: cmn – Mandarin Chinese; deu – Standard German; eng – English; esk – Northwest Alaska Inupiatun; grc – Koine Greek; mya – Burmese; tam – Tamil; qvw - Huaylla Wanca Quechua; vie – Vietnamese; xuo – Kuo; zul – Zulu.

This relationship between word order information and word structure information corresponds well with typological expectations. Highly synthetic languages like Inupiatun (ISO code: esk) or Quechua (qvw) have a higher level of word structure information and a lower level of word order information, while very analytic languages like Mandarin Chinese (cmn), Vietnamese (vie) or Kuo (xuo; an Mbum language of southern Chad) primarily convey grammatical information by the ordering of words (among them grammatical particles that correspond to inflectional morphology in more synthetic languages). On the other side, very analytical languages show a high(er) level of word order information and a low(er) level of word structure information. Languages like Koine Greek (the original language of the New Testament, grc), German (deu), or English (eng) mix both methods of conveying information; accordingly those languages tend to occupy intermediate spots on this spectrum.

To rule out the possibility of overfitting the data and to demonstrate the robustness of our results, Figure 2 shows that the resulting word order and word structure information rankings are strongly positively intra-correlated and strongly negatively inter-correlated. In terms of variance explained, if we know the word structure information ranking of Matthew, we can explain roughly $r^2 = .98^2 = 96\%$ of the variation in the word structure ranking in Acts and roughly $r^2 = -.74^2 = 54\%$ of the variation in its word order ranking.

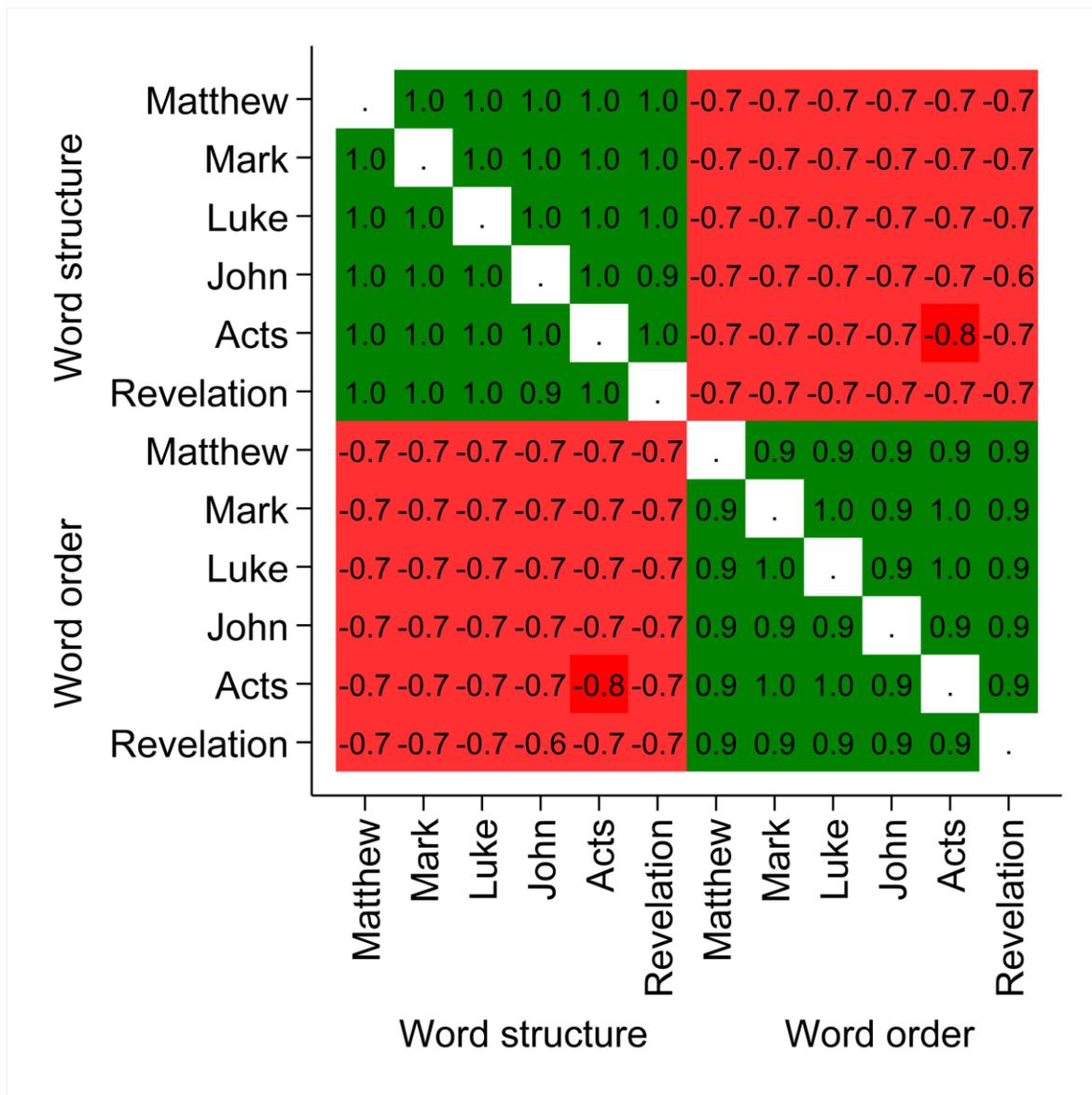

Figure 2. Spearman rank correlation matrix for all combinations of the six investigated books demonstrates that the word order and word structure information rankings have a strong positive intra-correlation and a strong negative inter-correlation.

Figure 3 shows that our results can also be interpreted from an evolutionary point of view. Here, we focus on Mark, since this is the only book for which we have available textual data in Old English. Old English, typologically classified as a synthetic language, relies on morphological structure to convey grammatical information. Modern English uses analytic constructions to mark grammatical relations. This corresponds well with the visible trend in

Figure 3, showing a substantial shift from Old English, with a high amount of word structure information to both Middle and Modern English, reducing the system of inflectional endings in favour of a stricter word order as indicated by a higher amount of word order information: "The Middle English evolution consists primarily in a shift towards more analytic structure, eventually approaching that of today's language which […] is close to isolating" [26].

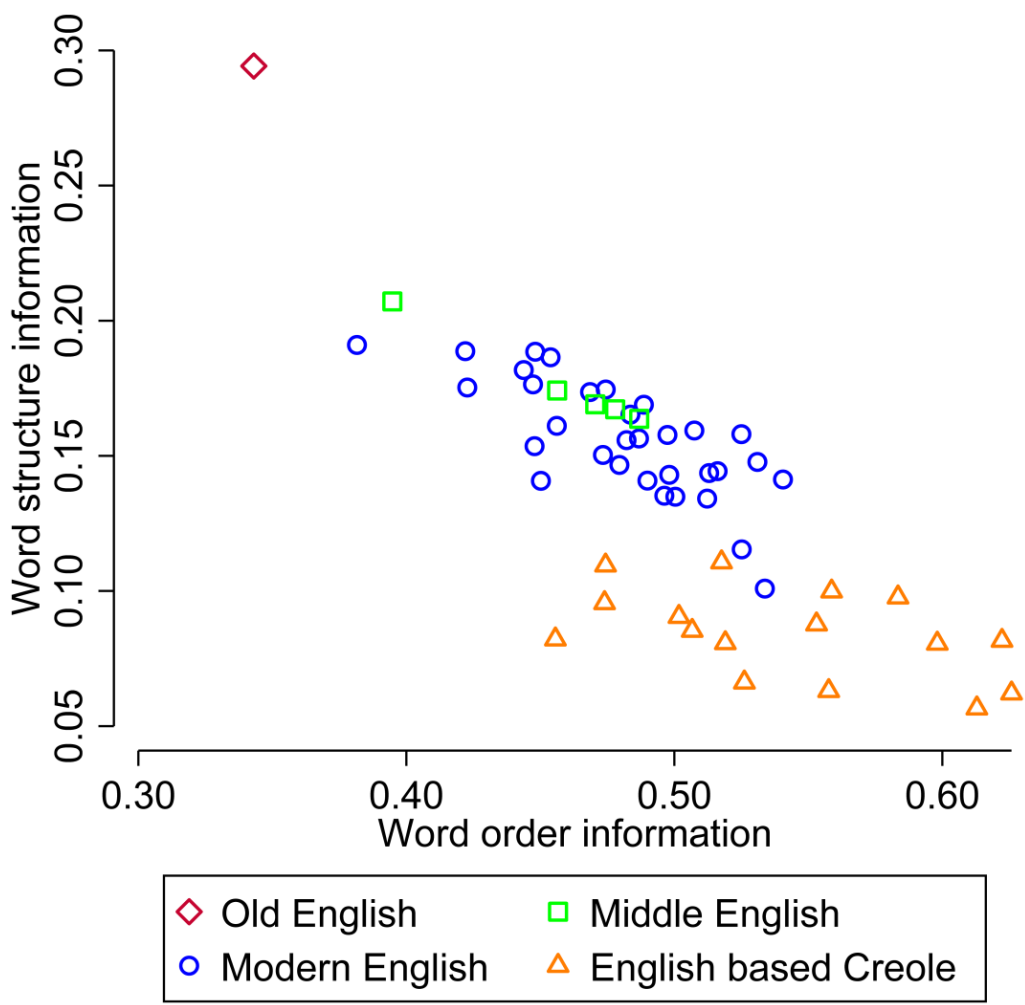

Figure 3.    Scatterplot depicting the relationship between word structure information and word order information for translations of the Gospel of Mark in English and English based Creole.

For example, with the loss of inflections in the period of Late Old English / Early Middle English, it became difficult to identify a genitive case when an article or a possessive adjective was followed by a noun phrase. This ambiguity problem was solved by replacing genitives with *of*-constructions in the period of Middle English [27] and thus creating a higher amount of word order information. Or put differently, if less information is carried within the word, more information has to be spread among words in order to communicate successfully. For the evolution of languages, this indicates that a change in one grammatical area can trigger a temporally subsequent change in another grammatical area. This is exactly what [28] shows for Icelandic, where changes within its words lead to changes in its syntax.

The classification of the English based Creole languages, which can be seen as "creative adaptions of natural languages" [2], also makes sense, as it indicates another substantial shift towards more analycity with very little reliance on inflectional morphemes compensated by a very strict word order to mark grammatical categories. It is worth pointing out that Figure 3 could also be used as a validation of the two key quantities estimated in this paper: the amount of word order information and the amount of word structure information measure how a given language encodes information, in this case, regarding grammatical functions.

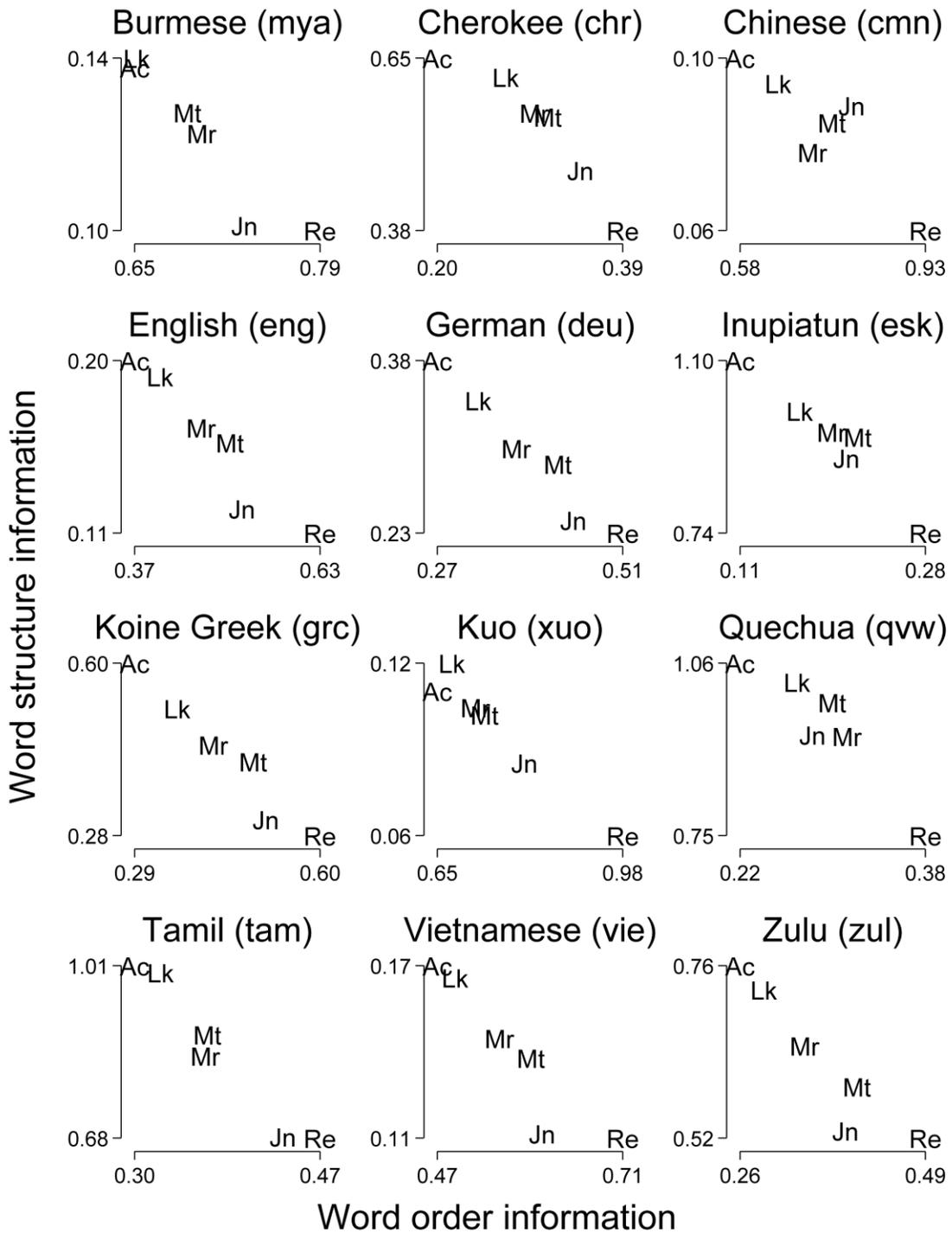

Figure 4. Inter-book trade-off for the six investigated books for twelve selected languages. Abbreviations: Ac – Acts; Jn – John; Lk – Luke; Mr – Mark; Mt – Matthew; Re – Revelation.

Remarkably, while Figure 2 clearly reveals that the inter-language word structure and word order rankings are strongly correlated, Figure 4 indicates that there also seems to be an inter-book trade-off in addition to the inter-language trade-off: the more informative the word order, the less informative the word structure and vice versa. If we calculate the Spearman correlation for the 6 investigated books, the median correlation for the 12 selected languages is $r_s = -.94$ (for all $N = 1,476$ translations, the median correlation is also $r_s = -.94$). If we compare the emerging pattern across languages, we find that the intra-language relationship is surprisingly similar: regardless of whether the languages are historically/geographically related or not, Revelation and John tend to have a higher word order information and a lower word structure information in relation to the other investigated books, and conversely for both Acts and Luke. Mark and Matthew tend to occupy intermediate positions. More precisely, if we calculate the correlations between all inter-book rankings for the selected languages, the median correlation is $r_s = .94$ in both cases. For the word structure ranking, the lowest correlation is between Mandarin Chinese and Kuo with $r_s = .71$ ($p = .068$). For the word order ranking, the smallest correlation is between Cherokee and Quechua with $r_s = .77$ ($p = .051$). In sum, these results suggest that while different languages occupy very different positions in the two-dimensional word order/word structure space indicating differences in the way grammatical information is encoded, content-related and stylistic features of the source material (here: the Koine Greek version) are encoded in very similar ways when they are translated into different languages.

To further explore this pattern in Figure 5, we used $N = 1,476$ translations for which we had available information for all six books. Here, we ranked the books, with a rank of 1 indicating that the corresponding book has the highest word order or word structure information of all six books of this translation. For each book, separate histograms visualizing the distribution of word order information ranks and word structure information ranks are depicted in the first two columns. The height of each bin represents the relative frequency of occurrence of the corresponding rank (in %). The matrix-plots (3rd columns) present bi-variate histograms, in which the colors of the cells represent the relative frequency of occurrence, with darker shades of gray representing a higher relative frequency (in %). In addition, numbers printed in each cell report relative frequencies rounded to the nearest integer. The emerging inter-book pattern in Figure 5 is remarkably stable across translations: Revelation and John tend to have a higher word order information and a lower word structure information in relation to the other investigated books, while the opposite applies to both Acts and Luke. Mark and Matthew tend to occupy intermediate positions across translations. While most contemporary scholars do not believe that the Revelation of John and the Gospel of John were written by the same person, there is a widespread consensus that the Book of Acts and the Gospel of Luke were written by one author [29]. In general, metaphors, symbolism and the repetition of key phrases are characteristic for the Book of Revelation [30], while some scholars describe the author of Luke-Acts as a reliable historian accurately recording historic events and geographic places [31]. Regarding the four Gospels, there is also agreement about the fact that Mark, Matthew and Luke are distinct from John, both in content and in style, with John containing more metaphors or allegories [32].

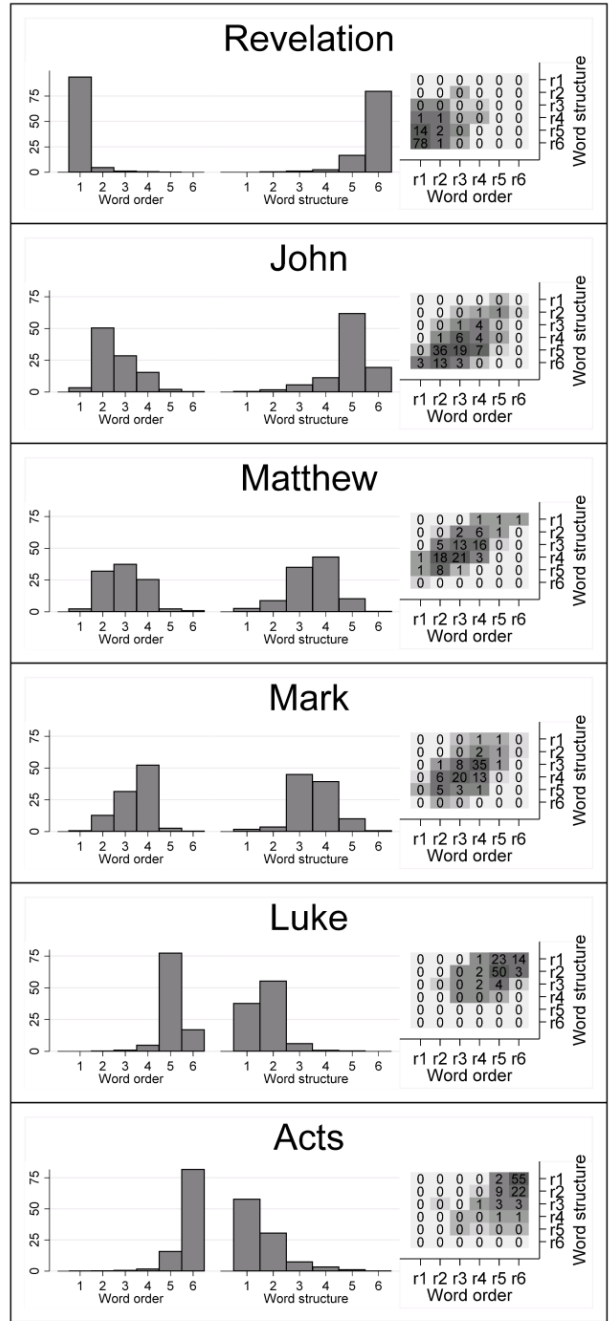

Figure 5. Word order information and word structure information rankings for the six investigated books. A rank of 1 indicates that the corresponding book has the highest word order or word structure information of all six books of the corresponding translation. Histograms visualizing the distribution of word order information ranks and word structure information ranks are depicted in the first two columns. The height of each bin represents the relative frequency of occurrence of the corresponding rank (in %). The matrix-plots (3rd columns) present bi-variate histograms, in which the colors of the cells represent the relative frequency of occurrence, with darker shades of gray representing a higher relative frequency (in %). In addition, numbers printed in each cell report relative frequencies rounded to the nearest integer.

Since our analysis has no liturgical goals, we conclude from a linguistic point of view that one interpretation of this result is that stylistic and content-related properties of the source material seem to be preserved when translated into different languages. Or put differently, if a book of the biblical canon shows more word order regularity and less word structure regularity than another book in one language, then this relationship between the books is likely to reappear in other languages. More work is needed to understand the linguistic correlates of style and/or content that lead to the emergence of this pattern.

## Discussion

In this paper we used a simple quantitative information-theoretic approach that is not restricted to a particular language or a particular writing system. Moreover, the approach is not motivated by a specific linguistic theory and is less subjective than other more traditional ad-hoc measures [8]. However, it is important to point out that such a numerical approach also has drawbacks: (i) It is completely based on the entropy estimation of symbolic sequences, the approach does not permit conclusions about specific word sequences (or alternative sequences) that can be used to encode a specific message. (ii) It does not reveal the precise underlying structures that are affected by the destruction of intra- and inter-word regularities [33]. (iii) Finding a definition of a word type that is valid across languages is harder than it might look [11]. For example, since we destroy the intra-lexical structure of distinct space-separated sequences, word types that span two strings separated by a blank are excluded from this procedure. This can be problematic, since the orthographical representation of compounding varies across languages, (e.g. English "data set" vs. German "Datensatz"). Addressing these problems and developing additional validation methods are clearly required in order to assess the accuracy of our approach.

With that in mind, we hope that we were able demonstrate that our results can be interpreted in an intuitively plausible and meaningful linguistic way [34]. We presented evidence that supports Zipf's principle of least effort in relation to the way natural languages encode grammatical information. In addition, we found that – despite differences in the way information is encoded – the inter-book trade-off regarding different texts of the biblical canon regarding the amount of both word order information and word structure information remains highly similar across translations into different languages. This result is both methodologically and theoretically remarkable. On the methodological side, it arose as a by-product of our study, something we did not expect when we conducted this study, which indicates the great potential of quantitative studies both for gaining empirical evidence for long-standing claims and for finding out new aspects about language and its statistical structure. On the theoretical side, the result suggests that basic stylistic and content-related properties of individual texts are preserved when they are translated into different languages.

Conversely, the stability of the relationship across books that differ with regard to style and content strengthens the evidence for the statistical trade-off between word order and word structure. Overall, this is our major scientific finding: there is an inverse relation between the amount of information contributed by the ordering of words and the amount of information contributed by the internal word structure: languages that rely more on word order to transmit grammatical information, rely less on intra-lexical regularities and vice versa.

## Acknowledgments

First and foremost, we would like to thank Michael Cysouw and the Parallel Bible Corpus team for compiling the Bible data and for making it available to us. Michael Cysouw also discussed the results with us and provided very insightful suggestions and comments. We thank Frank Michaelis for preparing the online interface. We also would like to thank Christian Bentz, Stefan Engelberg, Peter Fankhauser, and Marc Kupietz for their input and feedback. We are grateful to Katharina Ehret for making the Old English Bible data available to us. We thank Sarah Signer for proofreading. All remaining errors are ours.